\documentclass[11pt,a4paper]{article}
\usepackage{textcomp}
\usepackage{textcomp}
\usepackage[hyperref]{acl2021}
\usepackage{times}
\usepackage{latexsym}
\usepackage{graphicx} 
\usepackage{amssymb}
\usepackage[margin=1in]{geometry}
\usepackage{multirow}
\usepackage{graphicx}

\usepackage[leqno]{amsmath}
\usepackage{comment}
\aclfinalcopy

\begin{document}

\title{Meaning-infused grammar: Gradient Acceptability Shapes the Geometric Representations of Constructions in LLMs}

\author{Supantho Rakshit \\
  Dept of ECE / Princeton University \\
  \texttt{r.supantho@princeton.edu} \\\And
  Adele E. Goldberg \\
  Dept of Psychology / Princeton University\\
  \texttt{adele@princeton.edu} \\}
  
\maketitle

\date{}

\begin{abstract}
The usage-based constructionist (UCx) approach to language posits that language comprises a network of learned form-meaning pairings (\textit{constructions}) whose use is largely determined by their meanings or functions, requiring them to be graded and probabilistic. This study investigates whether the internal representations in Large Language Models (LLMs) reflect the proposed function-infused gradience. We analyze representations of the English Double Object (DO) and Prepositional Object (PO) constructions in Pythia-$1.4$B, using a dataset of $5000$ sentence pairs systematically varied by human-rated preference strength for DO or PO.   Geometric analyses show that the separability between the two constructions' representations, as measured by energy distance or Jensen-Shannon divergence, is systematically modulated by gradient preference strength, which depends on lexical and functional properties of sentences. That is, more prototypical exemplars of each construction occupy more distinct regions in activation space, compared to sentences that could have equally well have occured in either construction. These results provide evidence that LLMs learn rich, meaning-infused, graded representations of constructions and offer support for geometric measures for representations in LLMs.
\end{abstract}
\section{Introduction}

A central tenet of usage-based constructionist (UCx) approaches is that our knowledge of language consists of a structured inventory of constructions — conventionalized pairings of form \textit{and} function at varying levels of complexity and abstraction \cite{Goldberg2006}. The framework posits that language is learned from experience, with contexts and frequencies of use shaping dynamic "ConstructionNets" \cite{goldberg2024usage} in the minds of speakers.  Grammaticality is not a binary state but a continuum of acceptability, an observation supported by work in experimental and computational work on language \citep{francis2022gradient, gibson2013need, hu2024language}. 

Here we focus on two English constructions, the Double Object (DO) construction (e.g., \textit{She gave the boy the book}) and the Prepositional Object (PO) alternative (\textit{She gave the book to the boy}).  We build on a long-standing and widespread focus in linguistics on the combination of information structure and lexical factors that speakers use to choose between the DO and PO constructions: \citetext{e.g., \citealp{Bresnan2007, goldberg1995constructions,  green1974semantics, levin1993english, oehrle1976grammatical,  wasow2003post}}. In particular,  the recipient argument in the DO strongly tends to be already under discussion and expressed by a definite word (often a pronoun) or short phrase;  the transferred entity in a DO, on the other hand, is within the focus domain and is more often expressed by an indefinite noun phrase, which can be a longer phrase. These information structure properties partially emerge from the fact that the DO construction is used to convey real or metaphorical transfer to an animate entity, and animate entities are more likely to be topical in discourse (people often talk about people), while the transfered entity is more likely to be in the focus domain \citetext{for a degree of dialect variation see   \citealp{bresnan2009gradience}}.

The PO construction has been argued to be a subcase of a much broader ``caused-motion" construction \cite{goldberg1995constructions, goldberg2002surface} that can convey a change of location as well as transfer of possession (e.g., \textit{She kicked the ball to him/the wall}). This idea is supported by recent computational work offering a tool for analyzing word meanings in different contexts \cite{ranganathan2025semantic} using interpretable semantic features \cite{chronis2023method}. \citet{ranganathan2025semantic} report that the features associated with word embeddings vary systematically, depending on whether a given word appears in the DO or PO. In particular, features related to personhood are stronger when the same word, e.g., \textit{London}, is the recipient of the DO (\textit{e.g., She sent London the painting}), while features related to location are stronger when \textit{London} appears in the PO (e.g., \textit{She sent the painting to London}).  

Verbs' lexical biases also play a role in whether people prefer the DO or PO. The verb\textit{ give} is more common in the DO construction, and in fact \textit{give} accounts for roughly 40\% of all DO tokens \citetext{e.g., \citealp{goldberg2004learning}}. On the other hand, a set of Latinate (i.e., fancy-sounding) verbs resists the DO in favor of the PO \cite{gropen1989learnability, ambridge2012roles, goldberg2019explain}. For instance, the verbs \textit{transfer}, \textit{explain,} and \textit{donate} rarely occur in the DO, despite their highly compatible meanings; instead, each verb is biased toward the PO construction. Lexical biases can be quite particular and specific; for instance, the Latinate verb \textit{guarantee}, bucks the tendency for fancy-sounding words to resist appearing in the DO: \textit{Guarantee} strongly prefers the DO. Thus, a nuanced account of how such lexical factors are learned is required \citetext{e.g., \citealp{goldberg2011corpus, goldberg2019explain, ambridge2012roles}}. Indeed, computational work has found that the differences in information structure between the DO and PO are useful in LLMs' learning of lexical biases \citep{misra2024generating}.

As LLM representations are learned through exposure to natural language texts, there is an opportunity to investigate whether massive distributional learning can give rise to representations that reflect principles of the UCx approach.  Recent work has assessed how accurately LLMs can classify or distinguish argument structure constructions \citetext{e.g. \citealp{Huang2024, bonial2024constructing}}, 
but less is known about \textit{how} constructions are represented or their underlying geometry.  Our work addresses this gap by shifting the focus from classification accuracy to an analysis of underlying representational geometry. 

We hypothesize that representations of the DO and PO constructions should be more distinct to the extent that instances' typical lexical and functional properties are more prototypical instances of the respective constructions. We test this by asking whether collections of instances of the DO and PO that include typical functional features are more easily separable than collections of instances that are prototypical of neither, even though each sentence is unambiguous syntactically (either a PO or DO).

More specifically,  we ask: Does the geometric distinction between the representations of the DO and PO increase, as measured by either energy distance or Jensen-Shannon Divergence,  as the functional factors associated with DO and PO more closely align with their respective syntactic expressions?

Stimuli sentences come from the DAIS (Dative Alternation and Information Structure) dataset, which includes $5,000$ English pairs of DO and PO sentences \cite{Hawkins2020}.  Across DO/PO pairs, several factors are systematically varied along the dimensions recognized to distinguish the two constructions.  Here we use human preference strengths, also from DAIS, toward one or the other construction, to analyze the hidden states of Pythia-$1.4$B \cite{Biderman2023}.  

Both energy distance \cite{rizzo2016energy} and Jensen-Shannon divergence (JSD) \cite{fuglede2004jensen} are used to measure the separability of entire clouds of representations, at different layers of Pythia-$1.4$B. The preferred version (DO, PO, or either) of each sentence pair in the DAIS corpus was binned, according to the degree of preference toward the DO or PO, to be described in Methods and Results.

Results reveal a sophisticated geometric encoding of constructions in which lexical and functional factors improve the distinction between the DO and PO. In this way, LLM representations are consistent with key principles of the UCx approach, including gradiently distinguishable function-infused grammatical patterns, interpretable as clusters in geometric space.

\begin{figure*}[t]
    \centering
    \includegraphics[width=\textwidth]{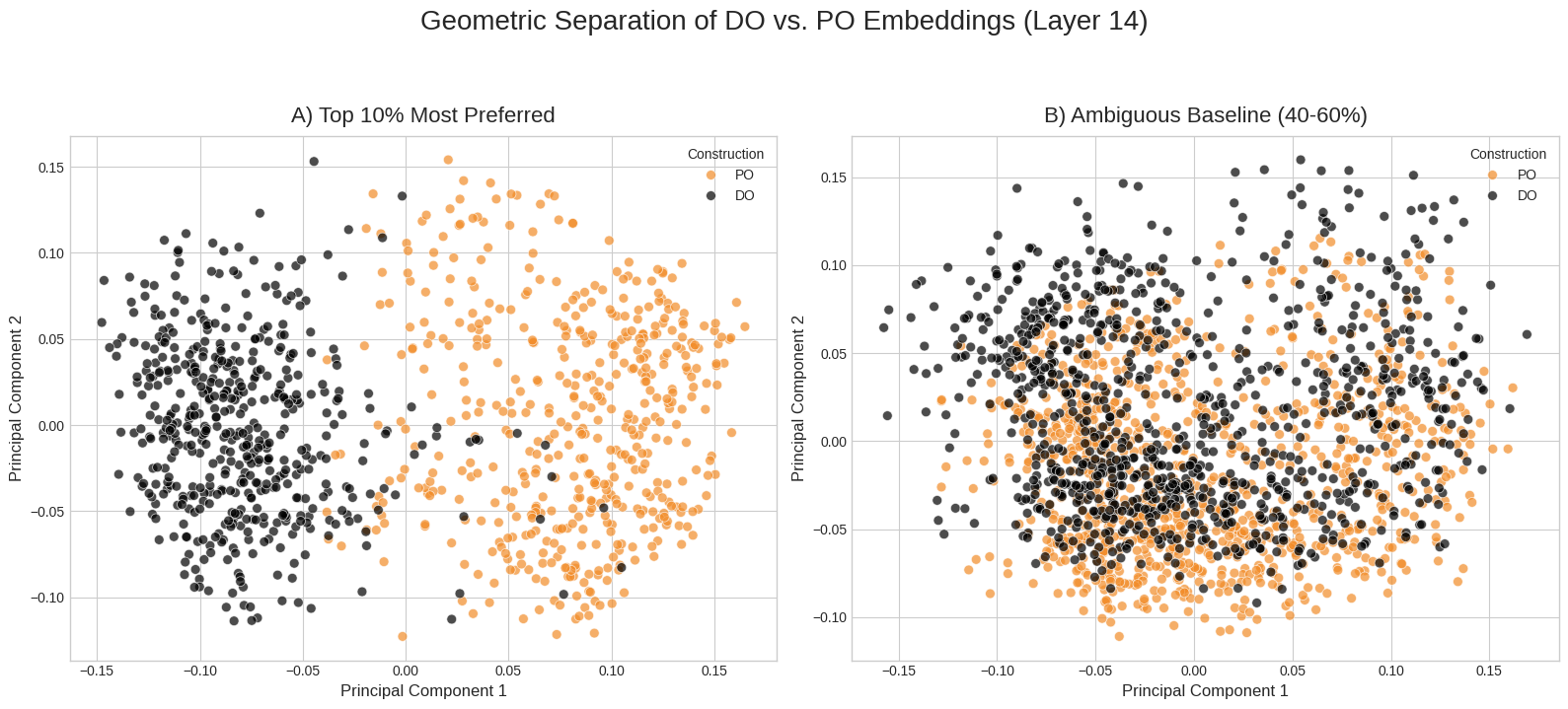}
    \caption{Projection into 2-dimensions of mean-pooled and normalized representations of the Double-Object (DO, in orange) and Prepositional Dative Object (PO, in black) constructions. Points represent sentences from the DAIS corpus, binned as follows:
    A) Instances of the DO and PO that are well-suited to the lexical and functional factors of their respective constructions as determined by human preferences.  
    B) Instances of the DO and PO, as determined by syntax only, as their lexical and functional properties do not favor either construction.}
    \label{fig:DO_PO_EDistnace}
\end{figure*}

\section{Methods}
\label{sec: Methods}
Dataset and Model: As noted, the DAIS dataset includes 5000 pairs of sentences, one in the DO and one in the PO, while systematically varying the length and definiteness each postverbal argument across pairs. Two hundred main verbs also vary across pairs, including verbs standardly treated as both `alternating' and 'non-alternating' \cite{levin1993english}. Importantly, DAIS also includes human ratings of how strongly they prefer one construction over the other, for each combination of verb and arguments. Participants used a slider to indicate a preference for the DO (one end) or PO (other end) or neither (midpoint) \cite{Hawkins2020}. 

We use these human preference ratings to partition sentences into five tiers based on the mean preference strength. We combined sentences from both ends of the scale to create 5 bins, ranging from: (1) the top $10 \%$ of sentences with the strongest preference for one or the other construction, to (5) those sentences judged to be in the middle of the scale (equally non-biased toward either construction). A sample  collection of DO sentences that vary from strong DO-bias (1) to little bias toward either DO or PO (5) are provided below: \\

\noindent
\begin{tabular}{@{}r@{\hspace{0.5em}}p{12cm}@{}}
\multicolumn{2}{@{}l}{DO biased} \\
\noindent
\textasciicircum & (1) Maria asked him some questions. \\
\textbar         & (2) Bob lobbed her a tennis ball. \\
\textbar         & (3) Juan shuttled the team something. \\
\textbar         & (4) Alice threw a woman a book. \\
\textbar         & (5) Michael took the woman the blanket. \\
\multicolumn{2}{@{}l}{DO or PO} \\
\end{tabular} 

\noindent An equal number of PO sentences were included in each of the same bins, correspondingly ranging from strongly PO-biased (1) to equi-biased (5), according to the human preferences in DAIS. 

From the publicly available pretrained Pythia-$1.4$B model, we extracted mean-pooled and normalized state representations for each sentence. We analyzed representations from all 24 layers, reducing them to $150$ principal components, which captured $88.01\%$ of the total variance (averaged across model layers). Common benchmarks suggest retaining components that explain 70\%–90\% of the total variance \cite{jolliffe2011pca}, and 
$88.01\%$ sits comfortably within this range, suggesting that a large majority of the structure in the data is retained, while a smaller portion that is more likely to reflect noise was discarded. We normalized the activations so that they all exist on a unit hypersphere $S^{149}$. Finally, we deployed the following analyses. \\

\noindent
\textbf{Preference strength} was treated as an ordinal variable with five levels: 1 (10\% most strongly biased) to 5  (10\% most equi-biased). That is, level (1) includes sentences that strongly preferred the DO and sentences that strongly preferred the PO, while level (5) included sentences that were roughly equi-biased toward either DO or PO. Bins were used rather than a continuous factor for visualization purposes.   \\

\noindent
To measure the separability in representational space for each tier of bias strength, we employed two different measures, Energy Distance and Jensen-Shannon divergence. \textbf{Energy distance}$, \mathcal{E}(X, Y)$, is a statistical distance between the probability distributions of two random vectors, $X$ and $Y$, in a metric space \cite{cramer1928composition}. It is simply based on the expected Euclidean distances between their elements. Given two samples from our PCA-reduced representations, $X=\left\{x_1, \ldots, x_m\right\}$ and $Y=\left\{y_1, \ldots, y_n\right\}$, where each $x_i, y_j \in \mathbb{R}^{150}$, the squared energy distance is estimated as:

\vspace{-0.8em}
\begin{equation}
\begin{split}
{\mathcal{E}}^2(X, Y) &= \frac{2}{mn} \sum_{i=1}^m \sum_{j=1}^n \left\|x_i - y_j\right\| - \\ \frac{1}{m^2} \sum_{i=1}^m \sum_{j=1}^m \left\|x_i - x_j\right\|
&\quad - \frac{1}{n^2} \sum_{i=1}^n \sum_{j=1}^n \left\|y_i - y_j\right\|
\end{split}
\nonumber
\end{equation}
\noindent 
where $\|\cdot\|$ is the Euclidean norm. Energy distance is zero if and only if the distributions are identical; it is sensitive to differences in both the location and the shape of distributions, making it a robust measure of overall geometric separation in the model's representation space.  We calculated the energy distance between the distributions of the constructions on $S^{149}$ layer by layer. 

A more sensitive measure of distributions is \textbf{Jensen-Shannon Divergence (JSD)}, which measures the relationship between distributions in high-dimensional space \cite{menendez1997jsd}. To use JSD, we first estimated the probability distributions of both constructions $P(v_{DO})$ and $Q(v_{PO})$, in the 150-dimensional PCA space. Following \citep{conklin2025information}, we
next generated a set of $k = 1000$ \textit{anchor vectors}, $A = a_1, a_2, \ldots, a_k$, by sampling from a uniform distribution on $S^{149}$.  The vector corresponding to each individual sentence $v$ was then assigned to the anchor vector nearest to it, based on cosine similarity. This effectively partitions the hypersphere into $k$ Voronoi cells. This technique, based on vector quantization, yields two discrete probability distributions, $\hat{P}$ and $\hat{Q}$, which are $k$-dimensional vectors where the $i$-th element represents the proportion of vectors from each set assigned to anchor $a_i$.  Jensen-Shannon divergence is then computed as:

\vspace{-0.8em}
\begin{equation}
JSD(\hat{P} \| \hat{Q}) = \frac{1}{2} D_{KL}(\hat{P} \| M) + \frac{1}{2} D_{KL}(\hat{Q} \| M)
\nonumber
\end{equation}
\noindent
where $M = \frac{1}{2}(\hat{P} + \hat{Q})$ and $D_{KL}$ is the Kullback-Leibler divergence. This method avoids information loss from projecting onto any single axis to offer a more holistic comparison of distributions \cite{conklin2025information}. Since the sampled anchor vectors are probabilistic, we averaged across $20$ random seeds to get stable JSD scores.

\section{Results}

\begin{figure}[t]
    \centering
    \includegraphics[width=\columnwidth]{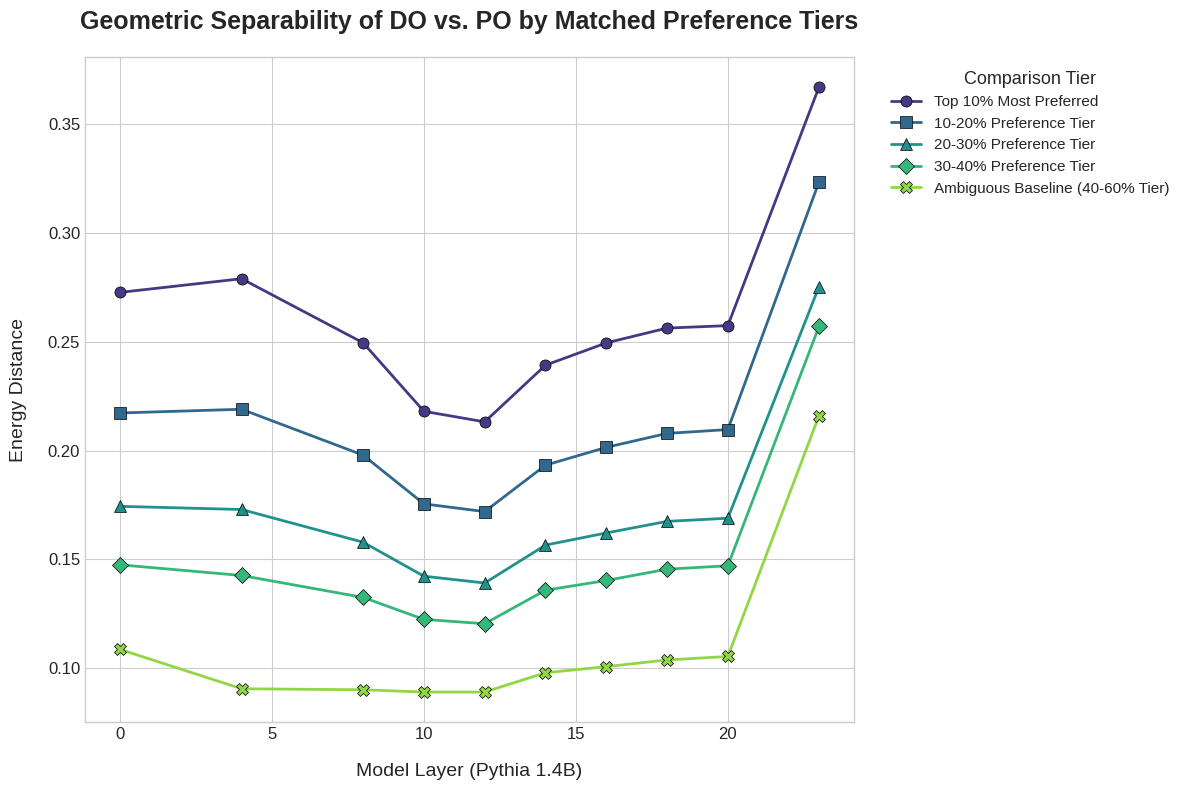}
    \caption{Layerwise Energy Distance between DO and PO representations, stratified by tiers binned by degree of bias. The plot shows a consistent ordering by bias, with more prototypical instances of the two constructions being more separable.}
    \label{fig:my_figure_energy_distance}
\end{figure}

\begin{figure}[t]
    \centering
    \includegraphics[width=\columnwidth]{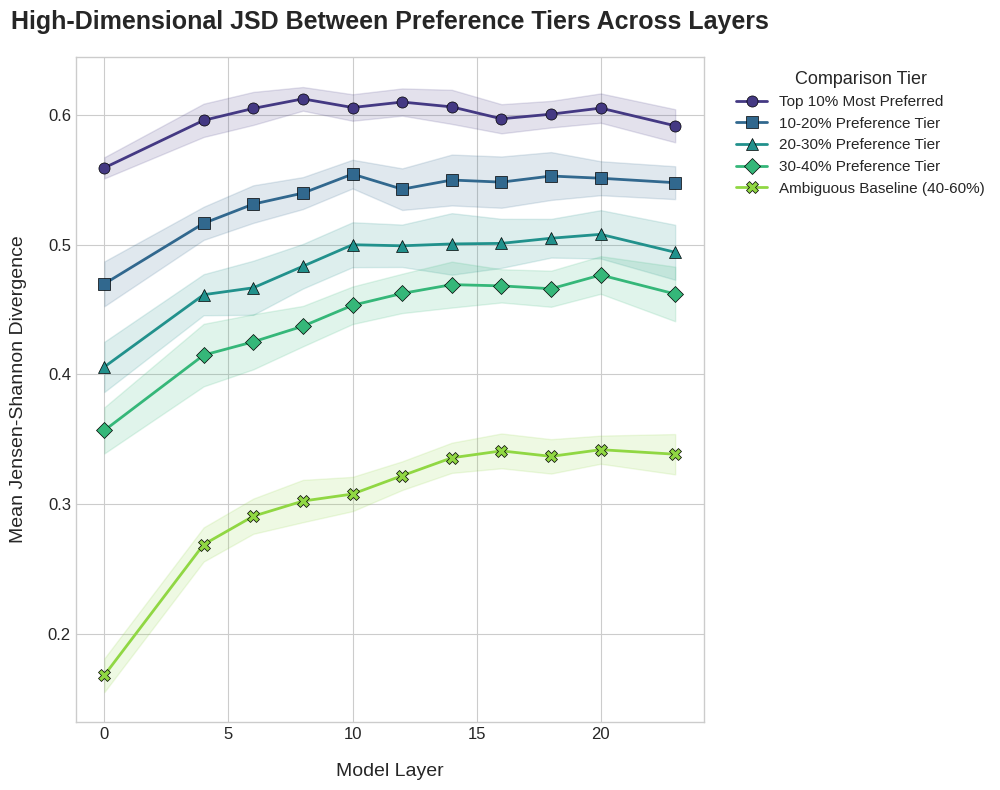}
    \caption{A high-dimensional measure of Distributional Separability (JSD) Across layers (with $k=1000$ anchor points, averaged over $20$ random seeds). Stratification is evident: tiers that include more biased sentences of either DO or PO are more separable into distinct constructions. This plot reinforces the finding from the energy distance analysis, though with different layer-wise dynamics.}
    \label{fig:my_figure_jsd}
\end{figure}

Our analysis reveals that graded bias strength for one construction over a paraphrase systematically shapes the geometry of construction representations across the model architectures. In particular, the model assigns representations that are more distinct when the constructions are more clearly differentiated, when instances are more strongly biased toward the construction used. This is the case for both energy distance and JSD, as each shows a clear and consistent stratification by the tiers of preference strength  (Figure ~\ref{fig:my_figure_energy_distance} and ~\ref{fig:my_figure_jsd}). At nearly every layer, the Top 10$\%$ strongest preference tier exhibits the greatest geometric distance, followed in order by the other tiers, down to the ambiguous baseline. As is clear in Figure 1, with sentence vectors projected onto two dimensions for visualization purposes,  DO and PO sentences that conform better to the DO or the PO, respectively (left panel), are more distinctive than sentences that could nearly as easily be paraphrased by the other construction (right panel).

Because energy distance and JSD are based on very different analyses, we cannot expect their qualitative patterning to align. In fact, energy distance follows a convex trajectory, slightly dipping in the mid-layers before rising sharply (Figure ~\ref{fig:my_figure_energy_distance})  In contrast, JSD shows divergence increasing sharply and remaining high (Figure ~\ref{fig:my_figure_jsd}). 
Yet the overall pattern showing more distinctiveness between DO vs. PO sentences when sentences that are more biased toward either varient compared with sentences that are relaitvely un-biased is evident according to both energy distance (Figure 3) and  JSD (Figure 2). We take this to indicate that the finding is robust and not an artifact of a single metric. 

\vspace{0.2cm}

\noindent \textbf{Scaling Analysis across Pythia Model Suite. } To test whether our findings are specific to the $1.4$B parameter model or reflect a more general property of transformer architectures, we replicated our full analysis pipeline — including the energy distance, high-dimensional JSD with correlation of mean cosine similarity tests — across the models in the Pythia suite (from $70$M to $6.9$B parameters). Results confirm that our central findings are robust across model scales. We consistently observe the geometric stratification by degree of bias so that preference strength remains a significant predictor of representational distance. A detailed report of these scaling law analyses, with code to generate plots, is in the \href{https://www.dropbox.com/scl/fo/vp0m6697wopwjcc5dklc4/ALHBF6n0AsdGTnmsNaqHNtw?rlkey=4b5lhi20xwmstbxkf942p39tu&st=jau2z6ka&dl=0}{supplementary materials}.     

\section{Discussion}
Our results provide compelling computational evidence for a core principle of the usage-based constructionist approach: that grammatical representations are graded and sensitive to semantic-pragmatic fit. The clear stratification in our geometric analyses demonstrates that LLMs develop representations whose geometric properties are highly consistent with the probabilistic, usage-based categories posited by the UCx approach. This geometric entanglement of form and function resonates with the core tenets of the UCx approach. The energy distance reflects the distinctiveness of the two constructions spatially in regions of the model's representation space. This extends previous work that has focused on the model's ability to classify constructions categorically \cite{Huang2024, bonial2024constructing} by showing more fine-grained, graded geometric structure, dependent on lexical and functional factors.  


Future work is needed to better understand the distinct qualitative patterns across layers when energy distance and JSD are compared. We note that the JSD measure, which is more nuanced but perhaps less intuitive, appears to distinguish the constructions particularly well: JSD is bounded between 0 and log(2) ($\approx 0.693$) \cite{lin1991divergence}, and the distinction between the constructions in the most biased tier approaches this limit at 0.6. Yet because the two metrics are based on quite different calculations, so we do not attempt to compare them directly.

\section{Conclusion and future directions}

The current work demonstrates that the geometry of an LLM's internal representations directly reflects the graded function-infused bias toward one or another linguistic construction, where biases are recognized to be conditioned on lexical and functional factors. We have shown that the model's representations of constructions are systematically organized by their distinctiveness. This work bridges the gap between the theoretical principles of the usage-based constructionist approach and the empirical realities of modern NLP, suggesting that LLMs learn a rich, dynamic, and meaning-infused model of grammar. Our findings open a promising new direction for future work; we are currently using these geometric insights to guide an investigation aimed at isolating the specific computational circuit(s) within the model that are responsible for encoding verb bias in the dative alternation, using tools like causal mediation analysis from Mechanistic Interpretability.

\section*{Acknowledgements}
We are grateful to the three anonymous reviewers for their insightful comments and constructive feedback, which significantly helped clarify and strengthen the arguments presented in this paper.

\bibliographystyle{acl_natbib}
\bibliography{references}


\end{document}